\documentclass[runningheads]{llncs}
\usepackage[T1]{fontenc}
\usepackage{graphicx}
\usepackage{amsmath,amssymb,bm}
\usepackage{algorithm, algorithmic}

\usepackage{xcolor}
\usepackage{ifthen}

\newcommand{\R}{\mathbb{R}}

\newcommand{\E}{\mathbb{E}}

\newcommand{\Tr}{\mathrm{Tr}}

\newcommand{\x}{\boldsymbol{x}}

\newcommand{\X}{\mathcal{X}}

\newcommand{\onemax}{O\textsc{ne}M\textsc{ax}}
\newcommand{\binval}{B\textsc{in}V\textsc{al}}
\newcommand{\leadingones}{L\textsc{eading}O\textsc{nes}}

\begin{document}
\title{Weight Adaptation for Improving Parallel Performance of Adaptive Stochastic Natural Gradient}
\titlerunning{Weight Adaptation for Improving Parallel Performance of ASNG}
%
\author{Yutaro Yamada \and
Kento Uchida \and
Shinichi Shirakawa}
%
\authorrunning{Y. Yamada et al.}
%
\institute{
Yokohama National University, Kanagawa, Japan\\
\email{yamada-yutaro-dw@ynu.jp, \\\{shirakawa-shinichi-bg, uchida-kento-fz\}@ynu.ac.jp}}

\maketitle              
\begin{abstract}
Probabilistic model-based evolutionary algorithms are promising for black-box optimization. Specifically, the adaptive stochastic natural gradient (ASNG) adaptively updates its learning rate, a typical hyperparameter in probabilistic model-based evolutionary algorithms, thereby realizing efficient and robust optimization. Although weight parameters are common hyperparameters, with the increasing demand for parallel evaluation of time-consuming tasks, it remains unclear how to set suitable weights for larger population sizes. In this paper, we propose Weight Adaptation ASNG (WA-ASNG), which incorporates a weight adaptation mechanism into ASNG. We calculated the estimated signal of the update direction from the accumulations of the natural gradient. Then, to maximize the signal, WA-ASNG adaptively updates its weight parameters by a gradient ascent over the optimization. While the learning rate adaptation plays a role in satisfying a sufficient condition for monotonic improvement of the expected objective value, the mechanism of weight adaptation is intended to maximize this improvement. The experimental results demonstrate that WA-ASNG outperforms PBIL and ASNG across various settings with population sizes ranging from 25 to 100 for binary optimization problems. Furthermore, WA-ASNG can perform efficiently in the presence of strong noise. 
Our code is available at \url{https://github.com/shiralab/WA-ASNG}.

\end{abstract}
\section{Introduction}
Probabilistic model-based evolutionary algorithms, such as estimation of distribution algorithms, utilize a probabilistic model for generating solutions and perform optimization by updating it based on the evaluation values.
This approach enables the derivation of a variety of efficient evolutionary algorithms, as exemplified by the population-based incremental learning (PBIL) \cite{pbil:improved} and covariance matrix adaptation evolutionary strategy (CMA-ES) \cite{hansen:cma} in black-box binary and continuous optimization, respectively.
Several probabilistic model-based evolutionary algorithms, including PBIL, can be derived from the information geometric optimization (IGO) \cite{ollivier:2017:jmlr} framework.
The IGO framework instantiates a specific IGO algorithm by applying a parametric probability distribution to the framework as the search distribution.
It updates the parameters of this distribution based on the natural gradient of the expected utility, where the utility function itself is defined by the quantiles of the objective function.
The utility function values are estimated from the sample rankings, and subsequently, a weight is assigned to each sample based on its rank.
The IGO framework is a form of stochastic natural gradient method, which is typically characterized by three hyperparameters: the population size, the learning rate, and the weight parameters.
Generally, the optimization performance depends on the settings of these parameters~\cite{lengler:2021,yang:2009}.

Recently, to reduce the cost of tuning problem-dependent hyperparameters, several techniques for adapting the learning rate or the population size have been proposed in the IGO-related algorithms.
For example, PSA-CMA-ES~\cite{psaCMA:2018} accumulates the update direction of the probability distribution into an evolution path, subsequently adapting its population size to keep the norm of this path constant.
On the other hand, adaptive stochastic natural gradient (ASNG)~\cite{asngnas} adaptively updates the learning rate such that a sufficient condition for monotonic improvement of expectations on the objective function is satisfied.
LRA-CMA-ES~\cite{LRACMA:2024} also adapts the learning rate to keep the signal-to-noise ratio (SNR) constant for the update directions of both the mean vector and the covariance matrix of the Gaussian distribution.

In contrast, research on weight parameters has primarily focused on theoretical analysis.
For the research on evolution strategies, the optimal weights have been derived for both the infinite-dimensional sphere function and a convex quadratic function~\cite{akimoto:2020:ecj,arnold:2005:foga}.
The default weights of CMA-ES are set similarly to these optimal weights~\cite{hansen:2014:principled}.
Moreover, for stochastic natural gradient methods, normalizing the weights to sum to zero is known to reduce the variance of the estimated natural gradient while its expected value remains unchanged~\cite{Evans:2000}.
This can justify the success of using negative weights in CMA-ES.
Similarly, it explains the effectiveness of the compact genetic algorithm (cGA)~\cite{cga}, where the update based on the worse solution corresponds to utilizing a negative weight in the context of stochastic natural gradient.

Due to the increase in computational power for parallel computation~\cite{openai:es:2017} and the emergence of high-throughput experimental platforms~\cite{parallel:2023}, the demand to utilize parallel evaluation has increased.
Probabilistic model-based evolutionary algorithms are well-suited for parallel computation, as a large population (sample) size effectively utilizes parallel evaluation.
However, it is unclear how to set suitable weights for larger population sizes, especially in a discrete domain such as binary optimization. Additionally, the appropriate weights will depend on the problem settings.
One approach to address this issue is to adaptively update the weight parameters according to the optimization progress.
Similar to adaptation mechanisms for the population size and the learning rate, this approach can also enhance optimization performance.

In this paper, we propose Weight Adaptation ASNG (WA-ASNG), which incorporates a weight adaptation mechanism into ASNG, a method that already features learning rate adaptation.
WA-ASNG estimates the signal of the update direction using the accumulations of the natural gradient.
Then, it adaptively updates the weight parameters by a gradient ascent to maximize the estimated signal.
Hence, WA-ASNG features two adaptation mechanisms: the learning rate adaptation inherited from ASNG, and our newly proposed weight adaptation.
Here, the former plays a role in approximately satisfying a sufficient condition for monotonic improvement of expectations on the objective function.
At the same time, the latter can be considered a mechanism to maximize this improvement.

We instantiate WA-ASNG by applying a Bernoulli distribution and evaluate it with population sizes ranging from 25 to 100 on binary optimization problems. The numerical experiments demonstrate that WA-ASNG outperforms PBIL and ASNG in various settings.
Furthermore, WA-ASNG shows better robustness than the variants of ASNG with different constant weights, even for noisy environments.

\section{Adaptive Stochastic Natural Gradient}
\subsection{Stochastic Natural Gradient Ascent}
Adaptive stochastic natural gradient (ASNG) \cite{asngnas} incorporates an adaptation mechanism of learning rate into a stochastic natural gradient ascent with stochastic relaxation of the objective function~\cite{ollivier:2017:jmlr}.
We consider an exponential family of distributions that has the expectation parameters $\bm{\theta} =\E_{p_{\bm{\theta}}}[T(\bm{x})]$ for the sufficient statistics $T:\X \to \R^{n_\theta}$ on the search space $\mathcal{X} \subseteq \mathbb{R}^N$.
Then, the probability density function is given by
\begin{align}
    p_{\bm{\theta}}(\bm{x}) = h(\bm{x}) \exp \left( (\eta(\bm{\theta}) )^\mathrm{T} T(\bm{x}) - \varphi(\bm{\theta}) \right)
    \enspace ,
\end{align}
where $\eta: \Theta \to \R^{n_\theta}$ is the natural parameter of this family and $\varphi:\Theta \to \R$ is the normalization factor.
We note that the exponential family includes several commonly used probability distributions in probabilistic model-based evolutionary algorithms, such as the Gaussian and Bernoulli distributions.
In the paper \cite{asngnas}, $h(\bm{x})$ is set as $h(\bm{x}) = 1$ for simplicity.

In ASNG, we maximize the expectation $J(\bm{\theta}) = \E_{p_{\bm{\theta}}}[f(\bm{x})]$ of the objective function $f:\X \to \R$ following the natural gradient method as
\begin{align}
    \bm{\theta}^{t+1} &= \bm{\theta}^{t} + \epsilon_\theta \tilde{\nabla}_{\bm{\theta}} J(\bm{\theta}^t) 
    \enspace ,
    \\
    \epsilon_\theta &= \delta_\theta / \| \tilde{\nabla}_{\bm{\theta}} J(\bm{\theta}^t) \|_{\mathbf{F}(\bm{\theta}^t)}
    \enspace ,
\end{align}
where $\mathbf{F}$ is the Fisher information matrix, $\delta_\theta > 0$ is the learning rate of ASNG, and $\tilde{\nabla}_{\bm{\theta}} = \mathbf{F}^{-1}(\bm{\theta}) \nabla_{\bm{\theta}}$.
Because $\tilde{\nabla}_{\bm{\theta}}J(\bm{\theta}^t)$ cannot be obtained analytically on black-box functions, we utilize Monte Carlo approximation to estimate it.
Under an exponential family of distributions with expectation parameters, we can calculate the natural gradient and the inverse of the Fisher information matrix as 
\begin{align}
    \mathbf{F}^{-1}(\bm{\theta}) &= \E_{p_{\bm{\theta}}}[ (T(\bm{x}) - \bm{\theta}) (T(\bm{x}) - \bm{\theta})^\mathrm{T} ]
    \qquad \text{and} \\
    \tilde{\nabla}_{\bm{\theta}} \ln p(\bm{x} \,|\, \bm{\theta}) &= T(\bm{x}) - \bm{\theta}
    \enspace .
    \label{eq:ng:est}
\end{align}
Therefore, using $\lambda_\theta$ samples, the approximation of $\tilde{\nabla}_{\bm{\theta}}J(\bm{\theta}^t)$ is given by
\begin{align}
    G_\theta(\bm{\theta}^t) = \frac{1}{\lambda_\theta} \sum^{\lambda_\theta}_{i=1} f(\bm{x}_i) (T(\bm{x}_i) - \bm{\theta}^t)
    \enspace .
\end{align}
Then, ASNG updates the parameter $\bm{\theta}$ with learning rate $\delta_\theta$ as
\begin{align}
    \bm{\theta}^{t+1} &= \bm{\theta}^{t} + \delta_\theta \frac{G_\theta(\bm{\theta}^t)}{\| G_\theta(\bm{\theta}^t) \|_{\mathbf{F}(\bm{\theta}^t)}}
    \enspace .
    \label{eq:asng:update}
\end{align}
When using a family of Bernoulli measures $p_{\bm{\theta}}(\bm{x}) = \prod_{i=1}^N p_{\theta_i}(x_i)$ with $p_{\theta_i}(x_i) = \theta_i^{x_i} (1-\theta_i)^{1-x_i}$ on the binary domain $\mathcal{X} = \{0,1\}^N$ (we simply call this \textit{Bernoulli distribution} in the paper), ASNG applies the margin correction to prevent premature convergence after the update of the distribution parameter as
\begin{align}
    \bm{\theta}^{t+1}_i \leftarrow \min\{ \max\{ \bm{\theta}^{t+1}_i, 1/N \}, 1 - 1/N \} \enspace.
    \label{eq:margin}
\end{align}

\subsection{Theoretical Background}
In the paper of ASNG \cite{asngnas}, the authors derive a sufficient condition for monotonic improvement of the expected objective function value from the following theorem.
\begin{theorem}[Theorem~4 in \cite{asngnas}]
\label{theorem:monotone}
Assume that $\min_{\x \in \X} f(\x) > 0$ and let $f^\ast = \max_{\x \in \X} f(\x)$.
For any $\epsilon > 0$, if
\begin{align}
    D_\theta(\bm{\theta}, \bm{\theta}^t + \epsilon \tilde{\nabla}_{\bm{\theta}} J(\bm{\theta}^t) )
    \leq \zeta D_\theta(\bm{\theta}^t + \epsilon \tilde{\nabla}_{\bm{\theta}} J(\bm{\theta}^t), \bm{\theta}^t)
    \label{eq:sufficient}
\end{align}
holds for some $\zeta > 0$, we have
\begin{align}
    \ln J(\bm{\theta}) - \ln J(\bm{\theta}^t)
    \geq \frac{1 - \zeta \epsilon f^\ast - \epsilon J(\bm{\theta}^t) }{\epsilon J(\bm{\theta}^t)} D_\theta(\bm{\theta}^t + \epsilon \tilde{\nabla}_{\bm{\theta}} J(\bm{\theta}^t), \bm{\theta}^t)
    \enspace ,
    \label{eq:improve}
\end{align}
where $D_\theta$ is KL-divergence.
In particular, if $\epsilon < ( \zeta f^\ast + J(\bm{\theta}^t) )^{-1}$ holds, we satisfy $J(\bm{\theta}) > J(\bm{\theta}^t)$.
\end{theorem}

In ASNG, the learning rate $\delta_\theta$ is adaptively updated such that $\bm{\theta} = \bm{\theta}^{t+1}$ satisfies \eqref{eq:sufficient}.
Under the assumption that $\delta_\theta$ is sufficiently small and the distribution does not change significantly over $\tau>0$ iterations with $\tau \propto 1 / \delta_\theta$, approximations of the natural gradient $G_\theta^{t+i}$ (i.e. update directions) are considered i.i.d. with $\E[G_\theta^{t+i}] = \tilde{\nabla}_{\bm{\theta}} J(\bm{\theta}^t)$ for $i=0, \cdots, \tau-1$.
Then, ASNG considers the relaxed condition of \eqref{eq:sufficient} obtained by replacing $\bm{\theta}$ with $\bm{\theta}^{t+\tau}$ and $\epsilon$ with $\tau \epsilon_\theta$.
In addition, by approximating the KL-divergence using the Fisher information norm, i.e., 
\begin{align}
    D_\theta(\bm{\theta}^{t+\tau}, \bm{\theta}^t + \tau \epsilon \tilde{\nabla}_{\bm{\theta}} J(\bm{\theta}^t) ) 
    &\approx \frac{ \epsilon_\theta^2 }{2} \cdot \left\| \sum_{i=0}^{\tau-1} G_\theta^{t+i} - \E[G_\theta^{t}] \right\|_{\mathbf{F}(\bm{\theta}^t)}^2
    \quad \text{and} \\
    D_\theta(\bm{\theta}^t + \tau \epsilon \tilde{\nabla}_{\bm{\theta}} J(\bm{\theta}^t), \bm{\theta}^t)
    &\approx \frac{ \tau^2 \epsilon_\theta^2 }{2} \cdot \| \E[G_\theta^{t}] \|_{\mathbf{F}(\bm{\theta}^t)}^2 \enspace,
\end{align}
and taking $\tau$ to $\infty$, the relaxed condition is transformed as
\begin{align}
    \Tr( \mathrm{Cov}[ G_\theta^{t} ] \mathbf{F}(\bm{\theta}^t) ) \leq \zeta \tau \| \E[ G_\theta^{t} ] \|_{\mathbf{F}(\bm{\theta}^t)}^2
    \enspace .
\end{align}
Therefore, it is necessary to maintain the value of SNR for the natural gradient at $\mathrm{\Omega}(\delta_\theta)$,
\begin{align}
    \frac{ \| \E[ G_\theta^{t} ] \|_{\mathbf{F}(\bm{\theta}^t)}^2 }{ \Tr( \mathrm{Cov}[ G_\theta^{t} ] \mathbf{F}(\bm{\theta}^t) ) } \geq \frac{1}{\zeta \tau} \in \mathrm{\Omega}(\delta_\theta)
    \enspace .
    \label{eq:snr:cond}
\end{align}

\subsection{Learning Rate Adaptation}
In order to satisfy \eqref{eq:snr:cond}, ASNG introduces the accumulation as follows:
\begin{align}
    \bm{s}^{t+1} &= (1 - \beta) \bm{s}^{t} + \sqrt{\beta (2 - \beta)} \mathbf{F}(\bm{\theta}^t)^{\frac{1}{2}} G_\theta^{t}
    \enspace ,\\
    \gamma^{t+1} &= (1 - \beta)^2 \gamma^{t} + \beta (2 - \beta) \| G_\theta^{t} \|_{\mathbf{F}(\bm{\theta}^t)}^2
    \enspace ,
\end{align}
where $\bm{s}^{0} = \bm{0}$, $\gamma^{0} = 0$, and $\beta = \delta_\theta / n_\theta^{\frac{1}{2}}$ is an accumulation factor.
If we assume that the distribution remains unchanged for $\tau$ iterations, we obtain the expectations of $\| \bm{s}^{t+\tau} \|^2$ and $\gamma^{t+\tau}$ as
\begin{align}
    \E[ \| \bm{s}^{t+\tau} \|^2 ] &\xrightarrow[\tau \to \infty]{}
    \frac{2 - \beta}{\beta} \| \E[ G_\theta^{t} ] \|_{\mathbf{F}(\bm{\theta}^t)}^2 + \Tr( \mathrm{Cov}[ G_\theta^{t} ] \mathbf{F}(\bm{\theta}^t) ) \label{eq:s2:expect}
    \enspace , \\
    \E[ \gamma^{t+\tau} ] &\xrightarrow[\tau \to \infty]{} \E[ \| G_\theta^{t} \|_{\mathbf{F}(\bm{\theta}^t)}^2 ] 
    \label{eq:gamma:expect}
    \enspace .
\end{align}
When replacing $\bm{s}^{t+\tau}$ and $\gamma^{t+\tau}$ with $\bm{s}^{t+1}$ and $\gamma^{t+1}$, repectively, and when satisfying $\| \bm{s}^{t+1} \|^2 / \gamma^{t+1} \geq \alpha$ for a constant $\alpha > 0$, the SNR value is approximately bounded as follows:
\begin{align}
    \frac{ \| \E[ G_\theta^{t} ] \|_{\mathbf{F}(\bm{\theta}^t)}^2 }{ \Tr( \mathrm{Cov}[ G_\theta^{t} ] \mathbf{F}(\bm{\theta}^t) ) } &\geq \frac{ \| \E[ G_\theta^{t} ] \|_{\mathbf{F}(\bm{\theta}^t)}^2 }{ \E[ \| G_\theta^{t} \|_{\mathbf{F}(\bm{\theta}^t)}^2 ] } \nonumber \\
    & \approx \frac{\beta}{2 - 2 \beta} \left( \frac{\| \bm{s}^{t+1} \|^2}{ \gamma^{t+1} } - 1 \right)
    \geq \frac{\beta (\alpha - 1)}{2 - 2 \beta} \in \mathrm{\Omega}(\delta_\theta)
    \enspace.
    \label{eq:snr_derive}
\end{align}
Therefore, ASNG updates the learning rate $\delta_\theta$ so as to satisfy $\| \bm{s}^{t+1} \|^2 / \gamma^{t+1} \approx \alpha$ by
\begin{align}
    \delta_\theta \leftarrow \delta_\theta \exp \left( \beta \left( \| \bm{s}^{t+1} \|^2 / \alpha - \gamma^{t+1} \right) \right)
    \enspace .
    \label{eq:asng:delta}
\end{align}

\subsection{Implementation of ASNG}
In order to improve stability, ASNG uses the following update rules in its implementation that accumulate the normalized value of the update direction $G_\theta^{t}$ as
\begin{align}
    \bm{s}^{t+1} &= (1 - \beta) \bm{s}^{t} + \sqrt{\beta (2 - \beta)} \frac{ \mathbf{F}(\bm{\theta}^t)^{\frac{1}{2}} G_\theta^{t}}{\| G_\theta^{t} \|_{\mathbf{F}(\bm{\theta}^t)}} 
    \enspace , \label{eq:accum:s} \\
    \gamma^{t+1} &= (1 - \beta)^2 \gamma^{t} + \beta (2 - \beta) 
    \enspace .
    \label{eq:accum:gamma}
\end{align}
Moreover, ASNG introduces a utility transformation based on the ranking of solutions.
Denoting the $i$-th best sample among $\lambda_\theta$ samples as $\x_{i:\lambda_\theta}$, ASNG transforms the objective value $f(\x_{i:\lambda_\theta})$ to the weight $w_i$ as follows:
\begin{align} \label{eq:asng_weight}
    w_i = \begin{cases}
    1 & \text{if}\quad i \leq \lceil \lambda_\theta / 4 \rceil \\
    0 & \text{if} \quad \lceil \lambda_\theta / 4 \rceil < i < \lfloor 3 \lambda_\theta / 4 \rfloor \\
    -1 & \text{if} \quad  \lfloor 3 \lambda_\theta / 4 \rfloor \leq i  \enspace .
    \end{cases}
\end{align}

\section{Weight Adaptation ASNG}
\begin{algorithm}[t]
\caption{Weight Adaptation ASNG}
\begin{algorithmic}[1]
\REQUIRE $t=0, \delta^{0} = 1, \eta_w = 1, L = N^{1.3}/\lambda_\theta, E=30$
\STATE Set an empty archive $\mathcal{A} = \emptyset$
\WHILE{termination condition is not met}
    \FOR{$i=1, \dots, \lambda_\theta$}
        \STATE Sample the candidate solutions as $\bm{x}_i \sim p(\bm{x}\,|\,\bm{\theta}^t)$
    \ENDFOR
    \STATE Evaluate the candidate solutions $\bm{x}_1, \dots, \bm{x}_{\lambda_\theta}$ on the objective function $f$
    \STATE Add the record $\{ ( \bm{x}_1, f(\x_1) ), \dots, ( \bm{x}_{\lambda_\theta}, f(\bm{x}_{\lambda_\theta}) ) \}$ to the archive $\mathcal{A}$
    \STATE Compute $G_\theta^t$ with the weight $\bm{w}$ using \eqref{eq:ng:est_weight}
    \STATE Update $\bm{\theta}^{t}$ using \eqref{eq:asng:update} (and apply margin correction~\eqref{eq:margin} if necessary)
    \STATE Update the accumulations $\bm{s}^{t}$ and $\gamma^{t}$ using \eqref{eq:accum:s} and \eqref{eq:accum:gamma}
    \STATE Update the learning rate $\delta_\theta^{t+1}$ and set the accumulation factor as $\beta = \delta_\theta^{t+1} / n_\theta^{\frac{1}{2}}$
    \IF{$t + 1 \equiv 0 \bmod L$}
        \FOR{$i=1, \cdots, E$}
        \STATE $\bm{s}_{i}^1 \leftarrow \bm{s}', \gamma_{i}^1 \leftarrow \gamma'$
            \FOR{$j=1, \cdots, L$}
                \STATE Compute $G_{\theta}^j$  using \eqref{eq:ng:est_weight} with the weight $\bm{w}$ and $j$-th record in $\mathcal{A}$
                \STATE Update the accumulations $\bm{s}^{j}_i$ and $\gamma^{j}_i$ using \eqref{eq:accum:s} and \eqref{eq:accum:gamma} with $G_\theta^j$.
            \ENDFOR
            \STATE Compute the signal as $S_i(\bm{w}) = \| \bm{s}_{i}^{L+1} \|^2 - \gamma_{i}^{L+1}$
            \STATE Update the weight as $\bm{w} \leftarrow \bm{w} + \eta_w \cdot \nabla_{\bm{w}} S_i(\bm{w})$
            \STATE Sort $\bm{w}$ in a descending order
            \STATE Update the learning rate for weight adaptation as $\eta_w \leftarrow 0.9 \times \eta_w$
        \ENDFOR
        \STATE Set $\bm{s}' \leftarrow \bm{s}_{i}^{L+1}, \gamma' \leftarrow \gamma_{i}^{L+1}, \mathcal{A} \leftarrow \emptyset, \eta_w\leftarrow1$
    \ENDIF
    \STATE $t \leftarrow t+1$
\ENDWHILE
\end{algorithmic}
\label{alg:proposed}
\end{algorithm}
The ASNG algorithm utilizes constant weight parameters defined in \eqref{eq:asng_weight}.
In this paper, we propose Weight Adaptation ASNG (WA-ASNG), which introduces a mechanism to adaptively update these weights to enhance search performance.
Specifically, WA-ASNG updates the weights so as to maximize the signal of the update direction $\| \E[ G_\theta^{t} ] \|_{\mathbf{F}(\bm{\theta}^t)}^2$.
Algorithm~\ref{alg:proposed} shows the update process of WA-ASNG.

\subsection{Relation between ASNG and Signal of Update Direction}
The learning rate adaptation mechanism of ASNG ensures high optimization performance by approximately satisfying \eqref{eq:sufficient}, the sufficient condition of Theorem~\ref{theorem:monotone}.
Furthermore, Equation~\eqref{eq:improve} provides a lower bound on the expected improvement of the objective function.
This bound is proportional to the KL-divergence $D_\theta(\bm{\theta}^t + \epsilon \E[ G_\theta^{t} ], \bm{\theta}^t)$ between the current probability distribution and the updated distribution using the true natural gradient direction.
Therefore, our approach is to maximize the expected improvement by maximizing this KL-divergence.
Assuming that the learning rate $\epsilon$ is sufficiently small, we can approximate the KL-divergence by Taylor expansion as
\begin{align}
    D_\theta(\bm{\theta}^t + \epsilon \E[ G_\theta^{t} ], \bm{\theta}^t) \approx \frac{\epsilon^2}{2} \cdot \| \E[ G_\theta^{t} ] \|_{\mathbf{F}(\bm{\theta}^t)}^2 \enspace.
\end{align}
Thus, WA-ASNG updates its weight parameters $\boldsymbol{w} = (w_1,\cdots, w_{\lambda_\theta})$ to maximize the signal of the update direction $\| \E[ G_\theta^{t} ] \|_{\mathbf{F}(\bm{\theta}^t)}^2$.
This update is performed while the learning rate adaptation mechanism continues to satisfy the condition in \eqref{eq:sufficient}, which is sufficient to guarantee a monotonic improvement of the expected objective function value.
We hypothesize that this leads to more efficient optimization by maintaining a larger improvement in the expected objective function value.

\subsection{Calculation of Signal}
Similar to the update rule of accumulations in ASNG, we assume that the learning rate is sufficiently small and that the distribution of the update direction remains unchanged.
In this case, by transforming \eqref{eq:snr_derive}, we can estimate the signal of the update direction as
\begin{align}
    \| \E[ G_\theta^{t} ] \|_{\mathbf{F}(\bm{\theta}^t)}^2
    \approx
    \frac{\beta}{2-2\beta} \Bigl( \| \bm{s}^{t+1} \|^2 - \gamma^{t+1} \Bigr)
    \enspace .
\end{align}
Because $\beta$ is constant, the maximization of the signal of the update direction is equivalent to that of $\| \bm{s}^{t+1} \|^2 - \gamma^{t+1}$.

\subsection{Weight Adaptation Mechanism}
WA-ASNG updates its weight parameters to maximize $\| \bm{s}^{t+1} \|^2 - \gamma^{t+1}$ by the gradient method.
Here, the accumulations $\bm{s}^{t+1}$ and $\gamma^{t+1}$ are functions of the estimated natural gradient $G_\theta^t = G_\theta(\bm{\theta}^t; \bm{w}^t)$.
This gradient is calculated using the current weights $\bm{w}^t = (w_1^t, \cdots, w_{\lambda_\theta}^t)$ as follows:
\begin{align}
    G_\theta(\bm{\theta}^t; \bm{w}^t) = \frac{1}{\lambda_\theta} \sum^{\lambda_\theta}_{i=1} w_i^t \cdot  (T(\bm{x}_{i:\lambda_\theta}) - \bm{\theta}^t)
    \enspace .
    \label{eq:ng:est_weight}
\end{align}
Thus, $S(\bm{w}^t) := \| \bm{s}^{t+1} \|^2 - \gamma^{t+1}$ is a function of the current weights $\bm{w}^t$.
Therefore, WA-ASNG updates its weight parameters in the direction of the gradient with respect to $S(\bm{w}^t)$ as
\begin{align}
    \bar{\bm{w}}^{t+1} = \bm{w}^t + \eta_w \cdot \nabla_{\bm{w}^t} S(\bm{w}^t)
    \enspace ,
    \label{eq:weight:adapt}
\end{align}
where $\eta_w > 0$ is the learning rate of the weight adaptation.
After this update, we obtain the new weights $\bm{w}^{t+1}$ by sorting the components of $\bar{\bm{w}}^{t+1}$ in descending order.

To stabilize the weight adaptation, the update of weights in \eqref{eq:weight:adapt} is performed every $L$ iterations in practice.
Here, $L \in \mathbb{N}$ is the weight update interval.
We calculate the gradient with respect to the weights that were used over the past $L$ iterations to obtain the accumulations $\bm{s}^{t+1}$ and $\gamma^{t+1}$.
The initial weight parameter $\bm{w}^{\mathrm{init}}$ is set to that used in ASNG, given by \eqref{eq:asng_weight}.
Additionally, the update process, including the gradient ascent update and sorting of the weight, is iteratively performed for $E \in \mathbb{N}$ epochs.
We note that the ASNG and our weight adaptation mechanism (with the accumulations in \eqref{eq:accum:s} and \eqref{eq:accum:gamma}) is invariant to the scaling transformation of the weight, including the initial weight.

\section{Experiment}
\begin{table}[t]
    \centering
    \caption{Definitions of the benchmark functions.}
    \begin{tabular}{c c l}
        \hline
        Name & \enspace & Definition \\
        \hline
        \hline
        \onemax & \enspace & $f(\x) = \sum_{k=1}^N x_k$ \\
        \binval & \enspace & $f(\x) = \sum_{k=1}^N 2^{k-1} x_k$ \\
        \leadingones & \enspace & $f(\x) = \sum_{k=1}^N \prod_{j=1}^k x_j$ \\
        \hline
    \end{tabular}
    \label{tab:def_benchmark}
\end{table}

\begin{figure}[t]
    \begin{center}
    \includegraphics[width=0.9\linewidth]{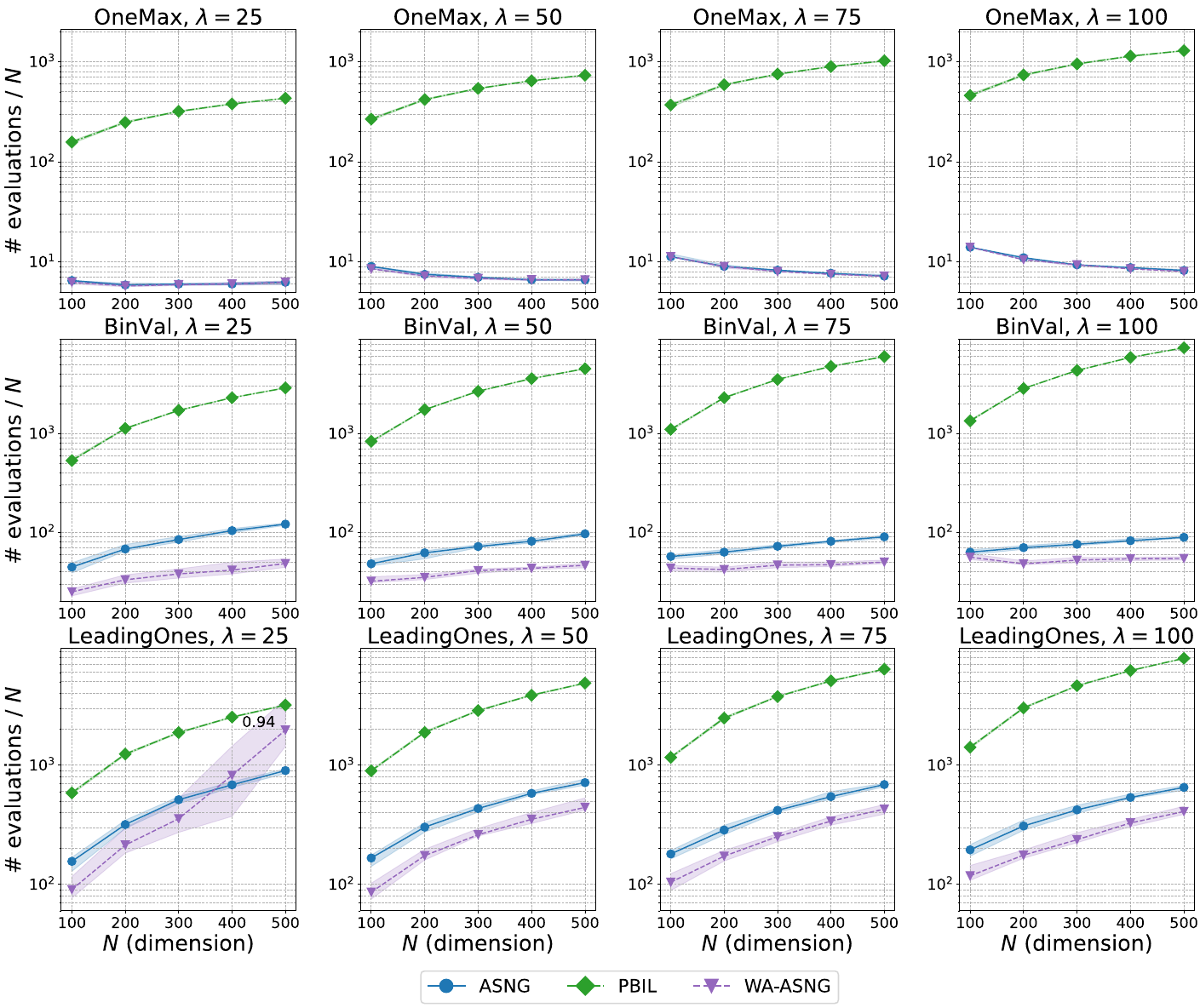}
    \end{center}
    \caption{The performance comparison between ASNG, PBIL and WA-ASNG (proposed). Each figure plots the medians and interquartile ranges of the number of evaluations over 31 trials. The success rate is shown near the marker if there exist one or more failed trials.}
    \label{fig:asng_pbil_waasng}
\end{figure}

\begin{figure}[t]
    \begin{center}
    \includegraphics[width=0.96\linewidth]{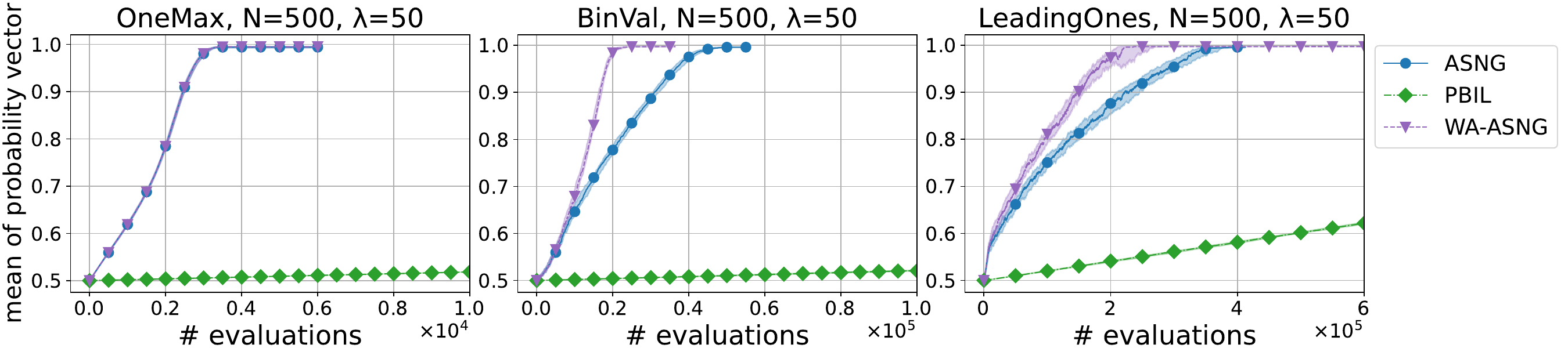}
    \end{center}
    \caption{The transitions of the mean of the elements in the probability vector in ASNG, PBIL, and WA-ASNG. We plot the medians and interquartile ranges over 31 trials.}
    \label{fig:probvec_asng_pbil_waasng}
\end{figure}

\begin{figure}[t]
    \begin{center}
    \includegraphics[width=0.65\linewidth]{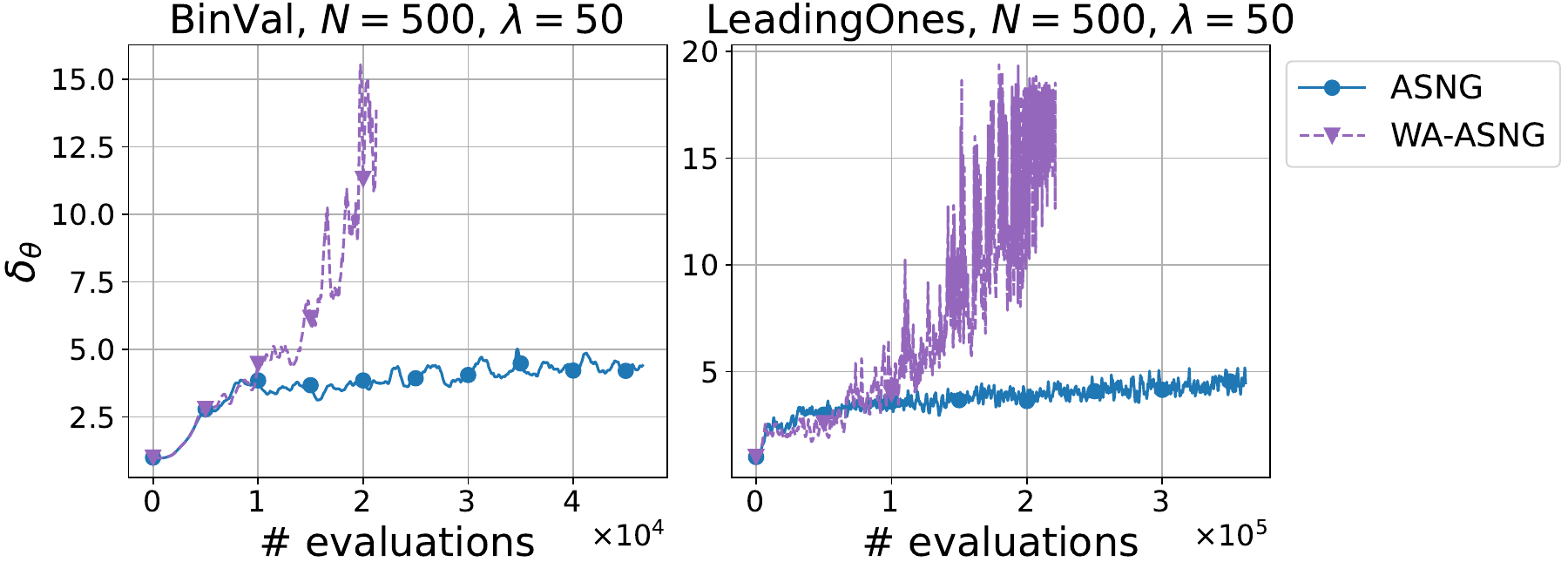}
    \end{center}
    \caption{The transitions of the learning rates $\delta_\theta$ in ASNG and WA-ASNG. We plot the result of typical trials on \binval{} and \leadingones{} for $N=500$ and $\lambda_\theta=100$.}
    \label{fig:delta_asng_waasng}
\end{figure}

\begin{figure}[t]
    \begin{center}
    \includegraphics[width=0.55\linewidth]{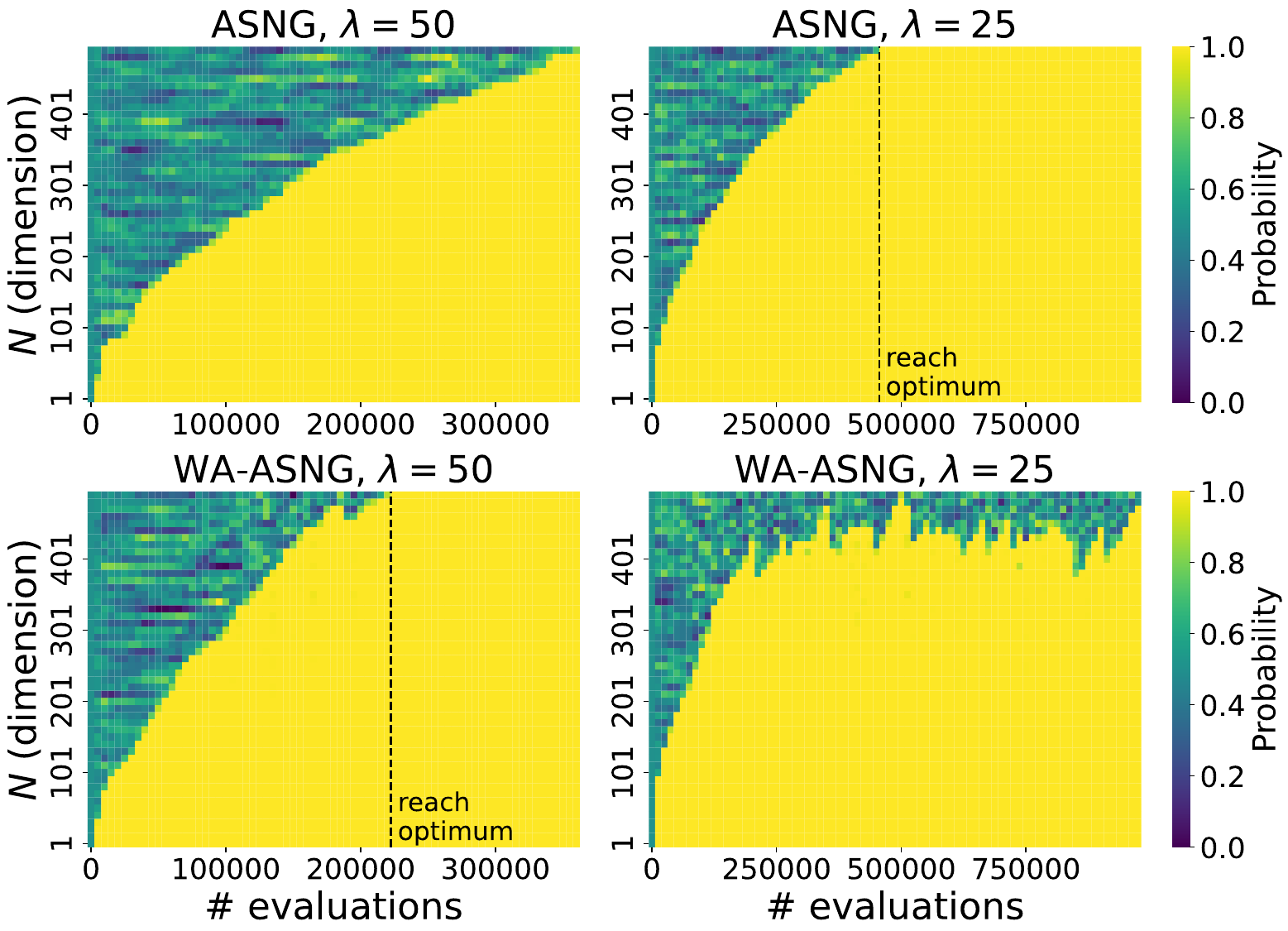}
    \end{center}
    \caption{
    The transitions of the elements in the probability vector in typical trials on \leadingones{} for $N=500$ with $\lambda_\theta=50$ (left) and $\lambda_\theta=25$ (right). 
    }
    \label{fig:prob_lo_asng_waasng}
\end{figure}

\begin{figure}[t]
    \begin{center}
    \includegraphics[width=0.9\linewidth]{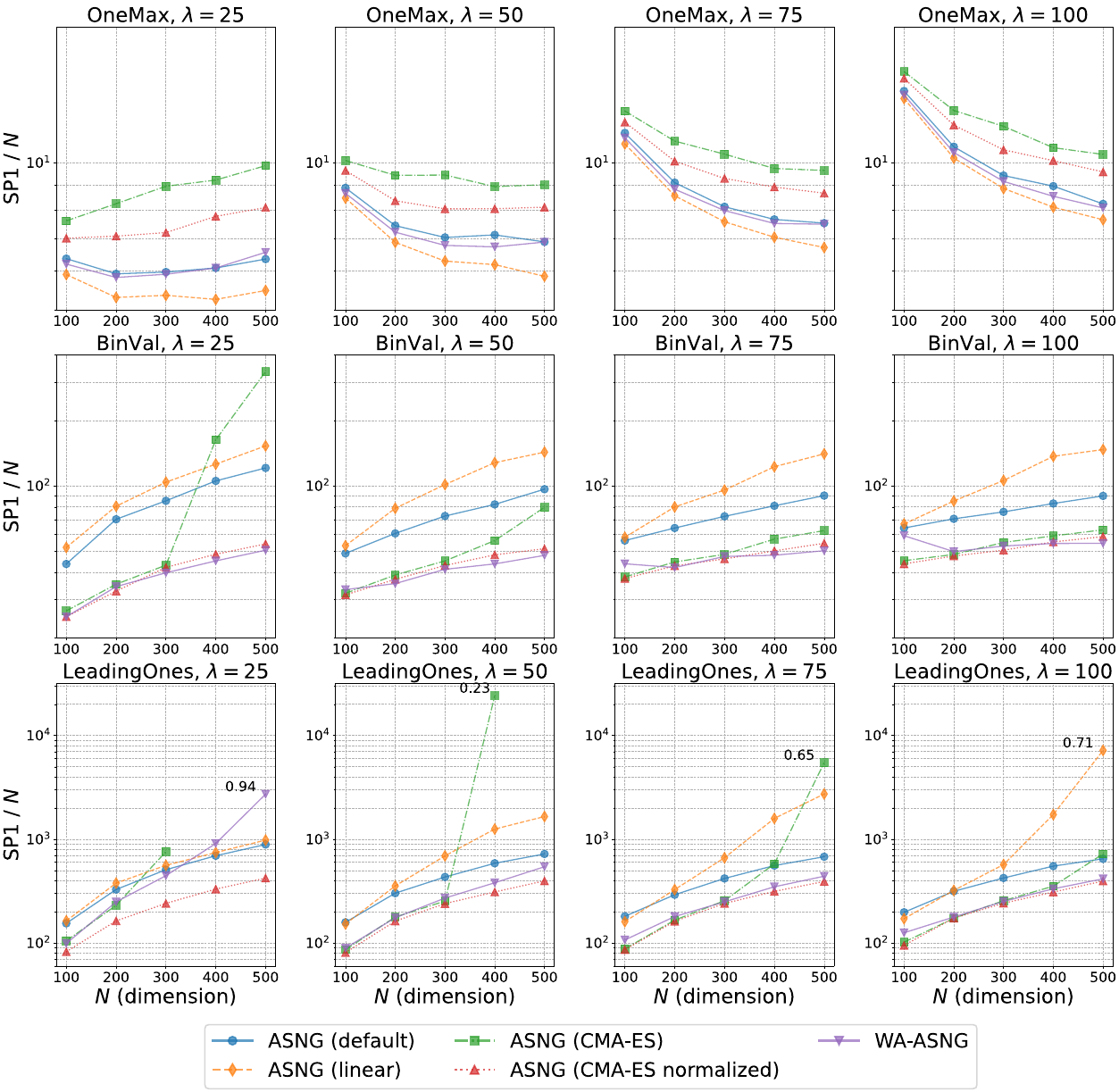}
    \end{center}
    \caption{
    The performance comparison between the variants of ASNG with different weight settings and WA-ASNG (proposed). Each figure plots SP1 value divided by the number of dimensions. The success rate is shown near the marker if there exist one or more failed trials. There is no marker when all trials failed.
    }
    \label{fig:asng_vars_waasng}
\end{figure}

\begin{figure}[t]
    \begin{center}
    \includegraphics[width=0.58\linewidth]{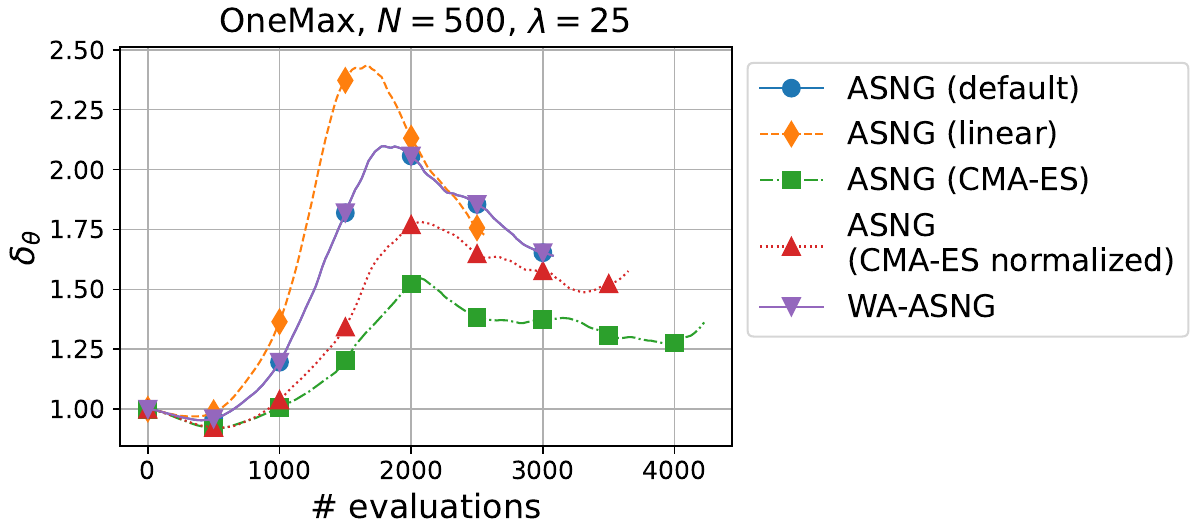}
    \end{center}
    \caption{
    The transitions of the learning rates $\delta_\theta$ in the ASNG with different weight parameters and in WA-ASNG (proposed). We plot the results of typical trials on \onemax{} for $N=500$ and $\lambda_\theta=25$.
    }
    \label{fig:delta_asng_vars_waasng}
\end{figure}

\begin{figure}[t]
    \begin{center}
    \includegraphics[width=0.8\linewidth]{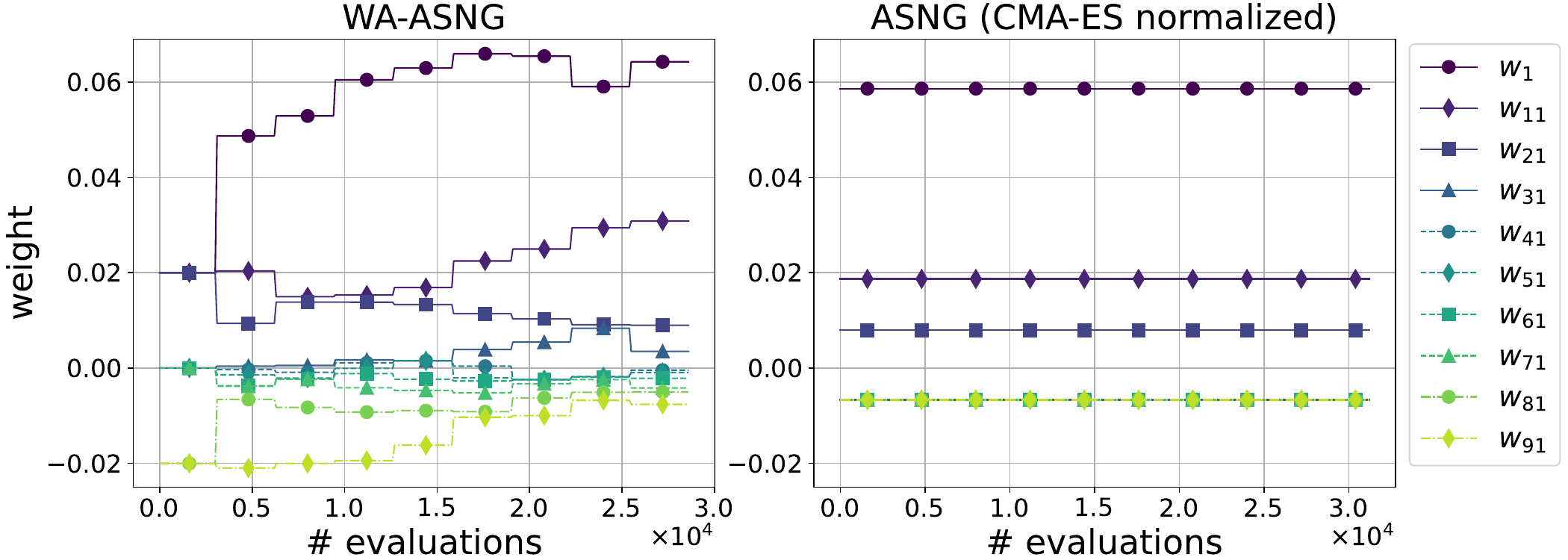}
    \end{center}
    \caption{
    The transitions of the weight parameters in WA-ASNG (proposed). We compare them with the ``CMA-ES normalized" weight. We plot the weight value for every 10 ranks in a typical trial on \binval{} with $N=500$ and $\lambda_\theta=100$.
    We plot the normalized weight so that the sum of their absolute values equals to that of ``CMA-ES normalized''.
    }
    \label{fig:weight_asng_vars_waasng}
\end{figure}

\begin{figure}[t]
    \begin{center}
    \includegraphics[width=0.58\linewidth]{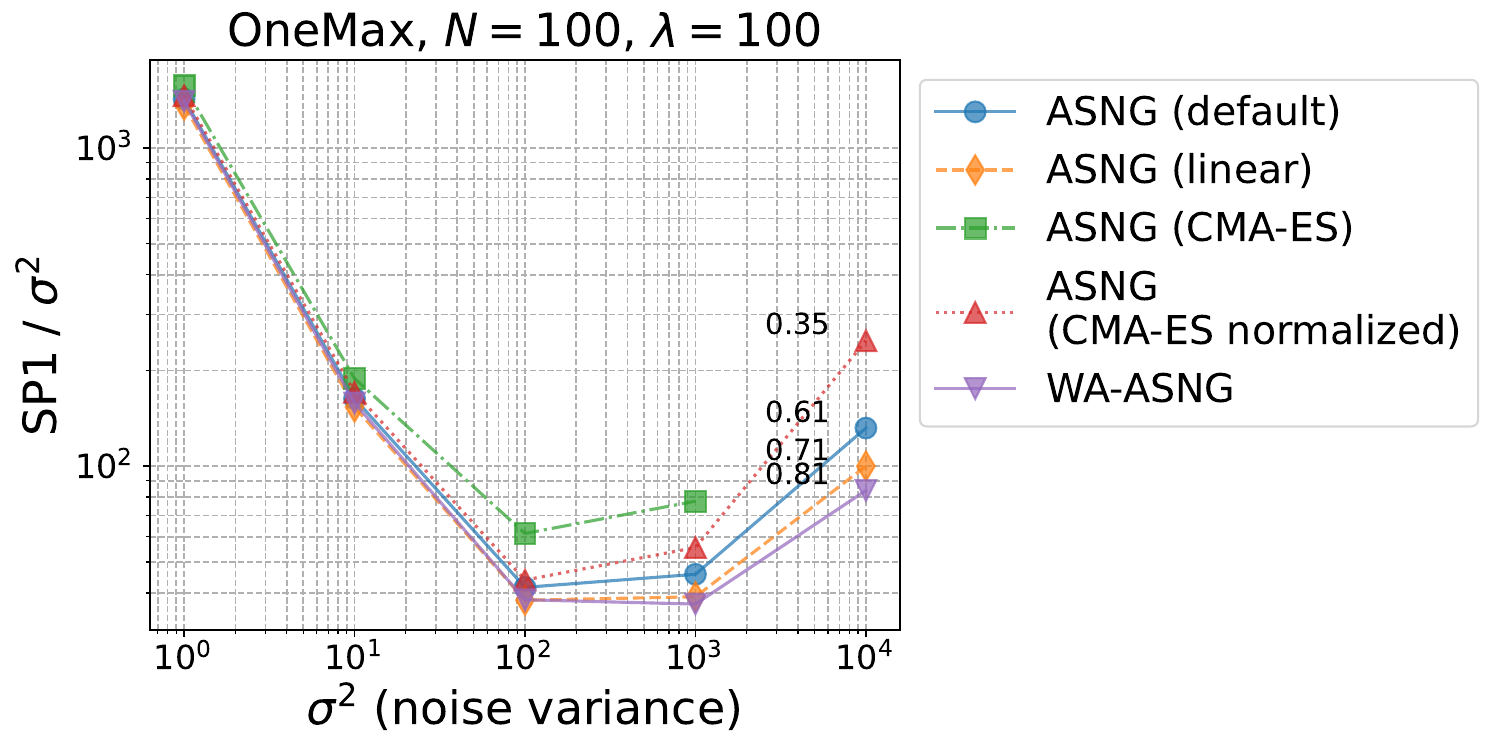}
    \end{center}
    \caption{The performance comparison between the variants of ASNG with different types for weight parameters and WA-ASNG (proposed). Each figure plots SP1 value divided by the noise variance on \onemax{} with Gaussian noise. The success rate is shown near the marker if there exist one or more failed trials. No marker indicates all trials failed.}
    \label{fig:noise_asng_vars_waasng}
\end{figure}

We evaluated the search performance of WA-ASNG with the Bernoulli distribution on binary optimization problems.

\subsection{Experimental Setting} \label{sec:exp_setting}
We utilized three benchmark functions whose search domain is the binary space $\mathcal{X} = \{ 0, 1 \}^N$: \onemax, \binval{} and \leadingones, commonly utilized for algorithm evaluations and analyses~\cite{dang:2019,asng-led}.
The definitions of these functions are shown in Table~\ref{tab:def_benchmark}.
In \onemax{}, all variables have the same contribution to the objective value, whereas variables in later dimensions have a greater influence in \binval{}.
In \leadingones{}, on the other hand, the influence of a variable is conditional on the values of all preceding variables, that fluctuate during optimization.
The optimal solution for these functions is $\x^*=(1, \cdots, 1)$, an $N$-dimensional vector whose variables are all one.
We varied the number of dimensions as $N=100, 200, 300, 400, 500$.

We evaluated the performance of ASNG and WA-ASNG by applying the multivariate Bernoulli distribution to them.
In this case, the number of dimensions of probabilistic distribution parameters $n_\theta$ is equal to $N$.
In addition, we compared the performance of PBIL with these methods.
We varied the population size as $\lambda_\theta=25,50,75,100$.
We set the hyperparameters of WA-ASNG as follows: the update interval was $L={N^{1.3}}/{\lambda_\theta}$, the initial learning rate for weight adaptation was $\eta_w = 1$, and the number of update epochs was $E=30$.
The learning rate of PBIL was set to $1/N$.
Other hyperparameters for ASNG and PBIL followed the default settings in \cite{asngnas} and \cite{pbil:improved}, respectively.

We ran 31 trials for each setting independently.
We considered a trial successful when the optimal solution was reached within the maximum number of function evaluations, which is set to $N \times 10^4$.

\subsection{Comparison with ASNG and PBIL}
First, we validated the search performance of WA-ASNG by comparing it with ASNG and PBIL.
Figure~\ref{fig:asng_pbil_waasng} shows the medians and the interquartile ranges of the number of evaluations divided by the number of dimensions on each benchmark function over 31 trials.
On \binval{}, WA-ASNG can find the optimal solution with the fewest evaluations for all settings of $\lambda_\theta$ and $N$.
Likewise, WA-ASNG outperformed ASNG and PBIL on \leadingones{} when $\lambda_\theta = 50, 75, 100$.

Figure~\ref{fig:probvec_asng_pbil_waasng} shows the transition of the probability vector on each problem.
It plots the medians and interquartile ranges of the mean of the probability vector over 31 trials.
As shown in the figure, WA-ASNG converged toward the optimal solution faster than ASNG and PBIL on \binval{} and \leadingones{}.
Figure~\ref{fig:delta_asng_waasng} illustrates how the learning rate $\delta_\theta$ for the probability vector changed during the optimization process in a typical trial on \binval{} and \leadingones{} with $N=500$ and $\lambda_\theta=50$.
WA-ASNG maintained a higher learning rate than that of ASNG, which facilitates faster convergence of the probability vector.

As shown in Figures~\ref{fig:asng_pbil_waasng} and \ref{fig:probvec_asng_pbil_waasng}, the performance of WA-ASNG is almost equal to that of ASNG on \onemax{}.
This is because the adaptation mechanism of ASNG is so efficient on simple problems like \onemax{} that the optimization terminates before the weight adaptation of WA-ASNG can take effect.

In addition, the performance of WA-ASNG deteriorates on \leadingones{} with $\lambda_\theta=25$ for $N=400$ and $N=500$.
As described in Section~\ref{sec:exp_setting}, the \leadingones{} problem requires an algorithm to optimize the dimensions sequentially, from the first to the last.
To investigate this behavior, Figure~{\ref{fig:prob_lo_asng_waasng}} shows the transition of the elements of the probability vector over the optimization in a typical trial on \leadingones{} with $N=500$.
The left column illustrates the case for $\lambda_\theta=50$, where WA-ASNG sequentially optimizes the probability vector faster than ASNG.
In contrast, the right column shows the case with $\lambda_\theta=25$ where ASNG performs better.
In this setting, WA-ASNG fails to optimize efficiently due to fluctuations in the probabilities of the latter components.
A possible reason is that WA-ASNG attempts to maximize the signal when the later parts of the probability vector change randomly, which leads to the fluctuation in the probability vector.

\subsection{Comparison with ASNG Using Various Weight Types}
We compared the performance of WA-ASNG with that of ASNG by setting its constant weights to the following different values:
\begin{itemize}
    \item default: the default weights of ASNG defined in \eqref{eq:asng_weight}. 
    \item linear: the linear weight on the interval $[-1, 1]$ as $w_i = 1 - \frac{2(i-1)}{\lambda_\theta-1}$. 
    \item CMA-ES: the default weight of the CMA-ES~\cite{hansen:cma}, one of the most successful methods in black-box continuous optimization, defined as
    \begin{align}
        w_i = w^\mathrm{cma}_i := \max \left( 0, \frac{\ln(\mu + 1) - \ln(i)}{\sum_{j=1}^{\mu} \ln(\mu + 1) - \ln(j)} \right)
        \quad \text{where} \enspace \mu = \lfloor \frac{\lambda_\theta}{2} \rfloor
        \enspace.
    \end{align}
    \item CMA-ES normalized: the normalized variant of $\bm{w}^\mathrm{cma}$ such that their sum is zero as $w_i = w^\mathrm{cma}_i - \frac{1}{\lambda_\theta}$. This weight was also used in \cite{xnes,ActiveCMA}.
\end{itemize}
All weights are assigned in descending order, i.e., $w_1 \geq w_2 \geq \cdots \geq w_{\lambda_\theta}$.

Figure~\ref{fig:asng_vars_waasng} shows the SP1 value divided by the number of dimensions for each benchmark function.
The SP1 is defined as the mean number of evaluations in successful trials, divided by the success rate.
The results with the various fixed weights show that the optimal weight setting strongly depends on the problem type, problem dimensionality, and population size.
For example, while ASNG~(linear) performed most efficiently on \onemax{}, ASNG~(CMA-ES~normalized) performed considerably better on \binval{}.
Figure~\ref{fig:delta_asng_vars_waasng} illustrates how the learning rates of the variants of ASNG changed in a typical trial on \onemax{} with $N=500$ and $\lambda_\theta=100$.
As shown in the figure, ASNG~(linear) maintains a higher learning rate during the optimization, which leads to better efficiency than the other variants.

Focusing on the proposed method, it exhibited robust performance on a wide range of problems compared to the other variants of ASNG.
The main exception is the \leadingones{} problem with $\lambda_\theta=25$, a setting where its performance deteriorates as previously mentioned.
Figure~\ref{fig:weight_asng_vars_waasng} shows how the weight adaptation of WA-ASNG worked on \binval{} with $N=500$ and  $\lambda_\theta=100$.
As the optimization proceeded, the weights of WA-ASNG changed significantly, eventually approaching the ``CMA-ES normalized'' weight.
This is noteworthy because the performance of ASNG (CMA-ES normalized) is nearly identical to that of WA-ASNG in this setting.

Interestingly, ASNG (CMA-ES normalized) was competitive with WA-ASNG on \binval{} and \leadingones{}.
In order to assess the robustness of WA-ASNG, we evaluated the performance of WA-ASNG and other variants of ASNG on \onemax{} for $N=100$ and $\lambda_\theta=100$ under evaluation noise.
Specifically, we added centered Gaussian noise with the variance $\sigma^2$ to the objective function value on \onemax{}.
We varied the variance as $\sigma^2 = 10^0, 10^1, 10^2, 10^3, 10^4$.
Similar to the setting of the previous experiment, we considered a trial successful when the optimal solution was found within the maximum number of function evaluations.
The maximum number of evaluations was set to $N \times 10^4$.
Figure~\ref{fig:noise_asng_vars_waasng} shows the SP1 value divided by the noise variance.
WA-ASNG achieved the lowest SP1 and the highest success rate, demonstrating its robust performance in noisy environments.
On the other hand, the performance of ASNG (default) and ASNG (CMA-ES normalized) deteriorated as the noise variance increased, despite their good performances on noiseless problems.

\section{Conclusion}
We proposed WA-ASNG, which introduces a weight adaptation mechanism to ASNG.
In our approach, we calculate the estimated signal of the update direction by accumulating the estimated natural gradient.
Then, to maximize this signal, we adaptively update the weight parameters by a gradient ascent according to the optimization progress.
While the learning rate adaptation inherited from ASNG satisfies a sufficient condition for monotonic improvement of the expected objective function value, the proposed weight adaptation mechanism attempts to maximize the rate of improvement.
Numerical experiments showed that WA-ASNG outperformed ASNG and PBIL on binary optimization problems.
We confirmed that the weight adaptation mechanism adaptively updated the weight values during optimization and maintained the higher learning rate of the probability distribution.
Furthermore, WA-ASNG achieved robust optimization performance on problems with noisy evaluations.

Despite the demonstrated effectiveness of the weight adaptation mechanism, we acknowledge several limitations that provide directions for future research.
First, our evaluation is primarily limited to the benchmark problems, \onemax{}, \binval{}, and \leadingones{}.
Although these benchmarks are commonly used, we need to validate the performance of WA-ASNG on a broader range of problems, including deceptive functions and categorical variable problems.
Second, the baseline method to which we introduced the weight adaptation mechanism was limited to ASNG.
Whether the proposed mechanism has transferability to other probabilistic model-based evolutionary algorithms, including those for continuous optimization, remains to be investigated.
We believe that addressing these limitations will be an important step toward developing more robust and widely applicable optimization algorithms.

\begin{credits}
\subsubsection{\ackname}
This study was partially funded by JSPS KAKENHI (JP24K20857, JP23K28156, and JP23H00491) and JST ACT-X (JPMJAX24C7).

\end{credits}

\bibliographystyle{splncs04}
\bibliography{reference}

\end{document}